\documentclass{llncs}
\usepackage{makeidx}  
\usepackage{graphicx}  
\begin{document}
\mainmatter              
\title{Training Deep Fourier Neural Networks To Fit Time-Series Data\thanks{The final publication is available at Springer via http://dx.doi.org/[insert DOI]}}
\titlerunning{Deep Fourier Neural Networks}
\author{Michael S. Gashler and Stephen C. Ashmore
}
\authorrunning{Michael S. Gashler and Stephen C. Ashmore} 
%
\tocauthor{Michael S. Gashler
}
\institute{University of Arkansas, Fayetteville AR, 72701, USA,\\
\email{\{mgashler, scashmor\}@uark.edu},\\ WWW home page:
\texttt{http://csce.uark.edu/$\sim$mgashler}
}

\maketitle              

\begin{abstract} 
We present a method for training a deep neural network containing sinusoidal activation
functions to fit to time-series data. Weights are initialized using a fast Fourier
transform, then trained with regularization to improve generalization. A simple dynamic
parameter tuning method is employed to adjust both the learning rate and regularization
term, such that stability and efficient training are both achieved. We show how deeper
layers can be utilized to model the observed sequence using a sparser set of sinusoid
units, and how non-uniform regularization can improve generalization by promoting the
shifting of weight toward simpler units. The method is demonstrated with time-series
problems to show that it leads to effective extrapolation of nonlinear trends.
\end{abstract}
\section{Introduction}\label{sec_introduction}

Finding an effective method for predicting nonlinear trends in time-series data is a
long-standing unsolved problem with numerous potential applications, including weather prediction,
market analysis, and control of dynamical systems. Fourier decompositions provide a mechanism
to make neural networks with sinusoidal activation functions fit to a training sequence
\cite{mingo2004fourier,tan2006fourier,zuo2009fourier}, but it is one thing to fit a curve to a
training sequence, and quite another to make it extrapolate effectively to predict
future nonlinear trends. We present a new method utilizing deep neural network training
techniques to transform a Fourier neural network into one that can facilitate practical
and effective extrapolation of future nonlinear trends.

Much of the work in machine learning thus far has focused on modeling static systems. These are
systems with behavior that depends only on a set of inputs, and do not change behavior over time.
One example of a static system is medical diagnosis. An electronic diagnostic tool could be
designed to evaluate the symptoms that a patient reports and attempt to label the patient with a
corresponding medical condition. Presumably, a correct diagnosis depends only on the reported
symptoms. Certainly, time plays a factor in the prevalence of medical conditions, but the change
is often either too slow, or too unpredictable to warrant consideration in the model. Thus, time
is effectively an irrelevant feature in this domain.

When time is a relevant feature, such as with climate patterns, market prices, population
dynamics, and control systems, one possible modeling approach is to simply treat time as
yet another input feature. For example, one could select a curve and adjust its coefficients
until the curve approximately fits with all the observations that have been made so far.
Predictions could then be made by feeding in an arbitrary time to compute a future prediction.

Nonlinear curve-fitting approaches tend to be very effective at interpolation, predicting
values among those for which it was trained, but they often struggle with extrapolation,
predicting values outside those for which it was trained. Because extrapolation requires
predicting in a region that is separated from all available samples, any superfluous
complexity in the model tends to render predictions very poor. Thus far, only very
simple models, such as linear regression, have been generally effective at extrapolating
trends in time-series data. Finding an effective general method for nonlinear extrapolation
remains an open challenge.

We use a deep artificial neural network to fit time-series data. Artificial neural networks
are not typically considered to be simple models. Indeed, a
neural network with only one hidden layer has been shown to be a universal function
approximator \cite{cybenko1989ann_universal_function_approximators}. Further, deep
neural networks, which have
multiple hidden layers, are used for their ability to fit to very complex functions.
For example, they have been very effective in the domain of visual
recognition \cite{ciresan2011committee,le2011building,krizhevsky2012imagenet,fan2014learning,taigman2014deepface}.
It would be intuitive to assume, therefore, that deep neural networks would be a poor
choice of model for extrapolation. However, our work demonstrates that a careful
approach to regularization can enable complex models to extrapolate effectively,
even with nonlinear trends.

This document is layed out as follows: Section~\ref{sec_intuition} gives an intuitive-level description
of our approach for fitting curves to time-series data. Section~\ref{sec_related} summarizes
related works. Section~\ref{sec_technical} gives a technical description of our approach.
Section~\ref{sec_validation} reports results that validate our approach. Finally,
Section~\ref{sec_contributions} summarizes the contributions of this work.

\section{Intuitive-level Algorithm Description}\label{sec_intuition}

\begin{figure}[!tb]
	\begin{center}
		\includegraphics[width=3.2in]{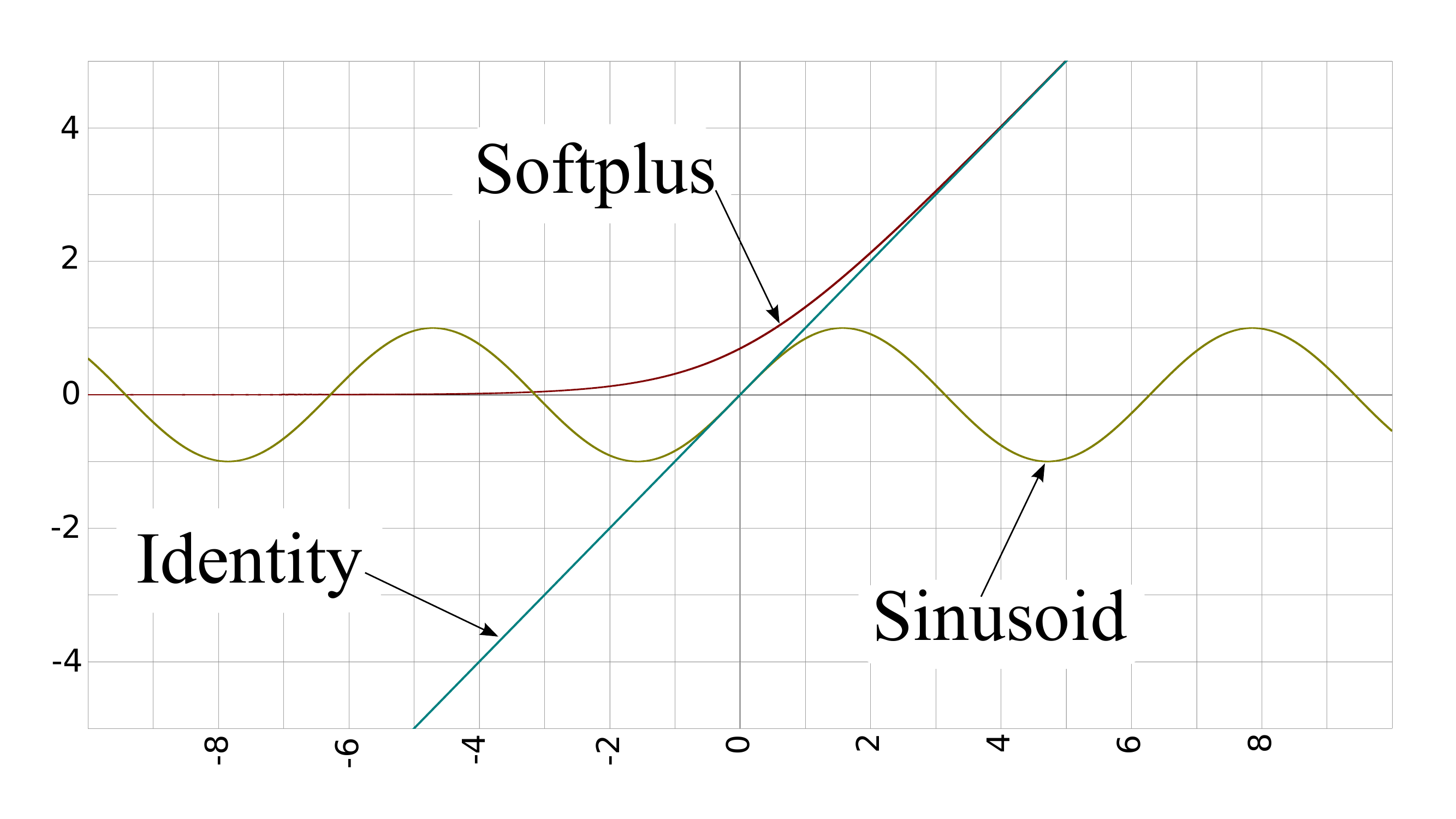}
		\caption{Each unit in our artificial neural network uses one of three
				activation functions, sinusoid, softplus, or identity.}
		\label{fig_activations}
	\end{center}
\end{figure}

Our approach uses a deep artificial neural network with a mixture of activation
functions. We train the network using stochastic gradient descent \cite{wilson2003general}.
Each unit in the artificial neural network uses one of three activation functions,
illustrated in Figure~\ref{fig_activations},
sinusoid: $f(x) = sin(x)$,
softplus: $f(x) = \log_e(1 + e^x)$, or
identity: $f(x) = x$.
Using the ``identity" activation function creates a linear unit, which is only capable of
modeling linear components in the data.
Nonlinear components in the data require a nonlinear activation
function. The softplus units enable the network to fit to non-repeating nonlinearities
in the training sequence. The sinusoid units enable the network to fit to repeating nonlinearities
in the data.

\begin{figure}[!tb]
	\begin{center}
		\includegraphics[width=2.1in]{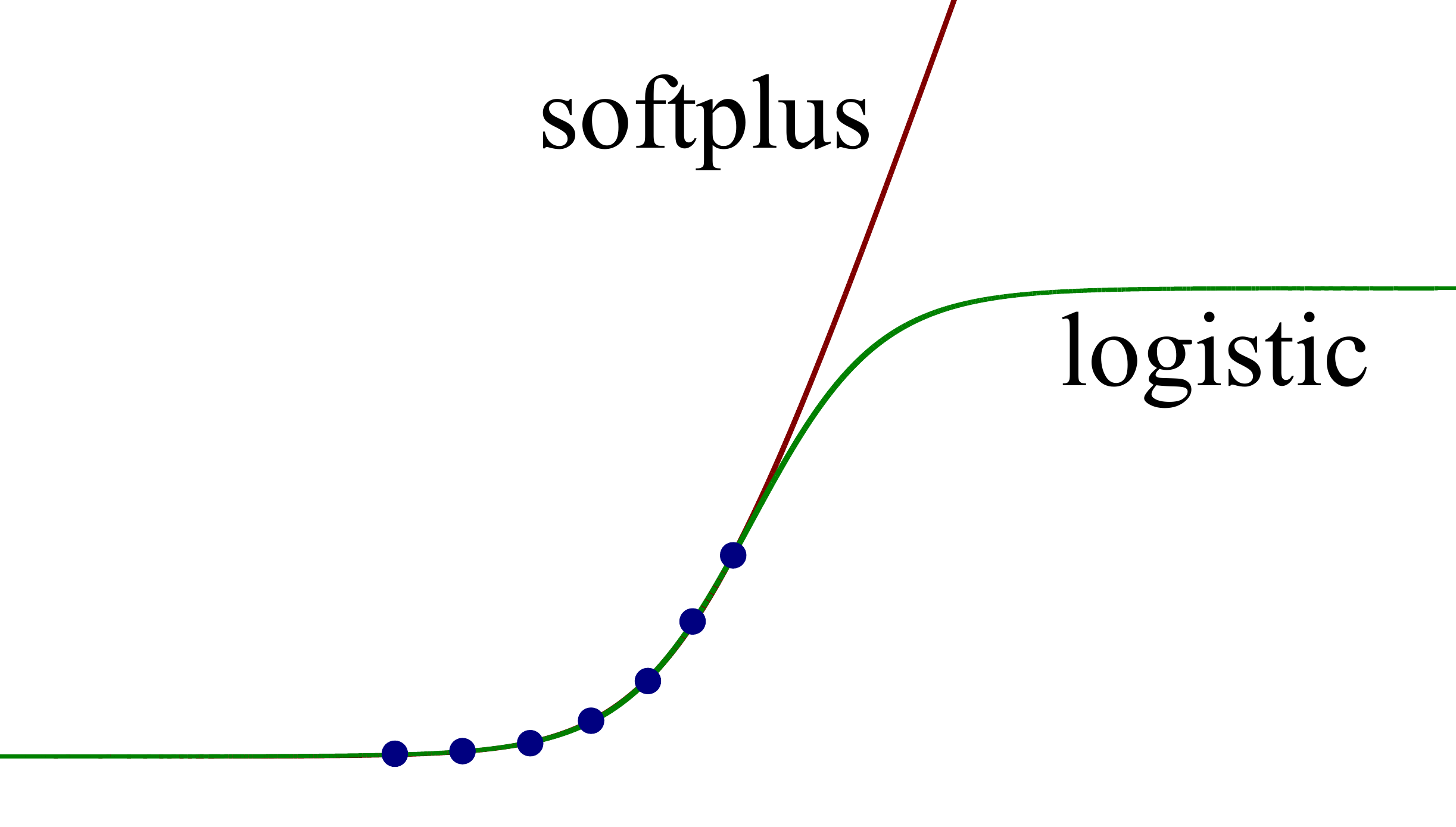}
		\caption{We used 8 datapoints to represent a curve with a single bend, and fitted two
				models to it. The green curve is a model consisting of a single logistic unit
				that feeds into a linear unit (to enable it to fit to arbitrary values), and the
				red curve is a model consisting of a single softplus unit that feeds into a
				linear unit. Both models can approximate this training data closely, but the logistic
				model exhibits a superfluous bend that is not directed by any property in the data.
				Therefore, to be consistent with Occam's razor, softplus units are a better choice.}
		\label{fig_superfluous_bends}
	\end{center}
\end{figure}

\begin{figure}[!tb]
	\begin{center}
		\includegraphics[width=3in]{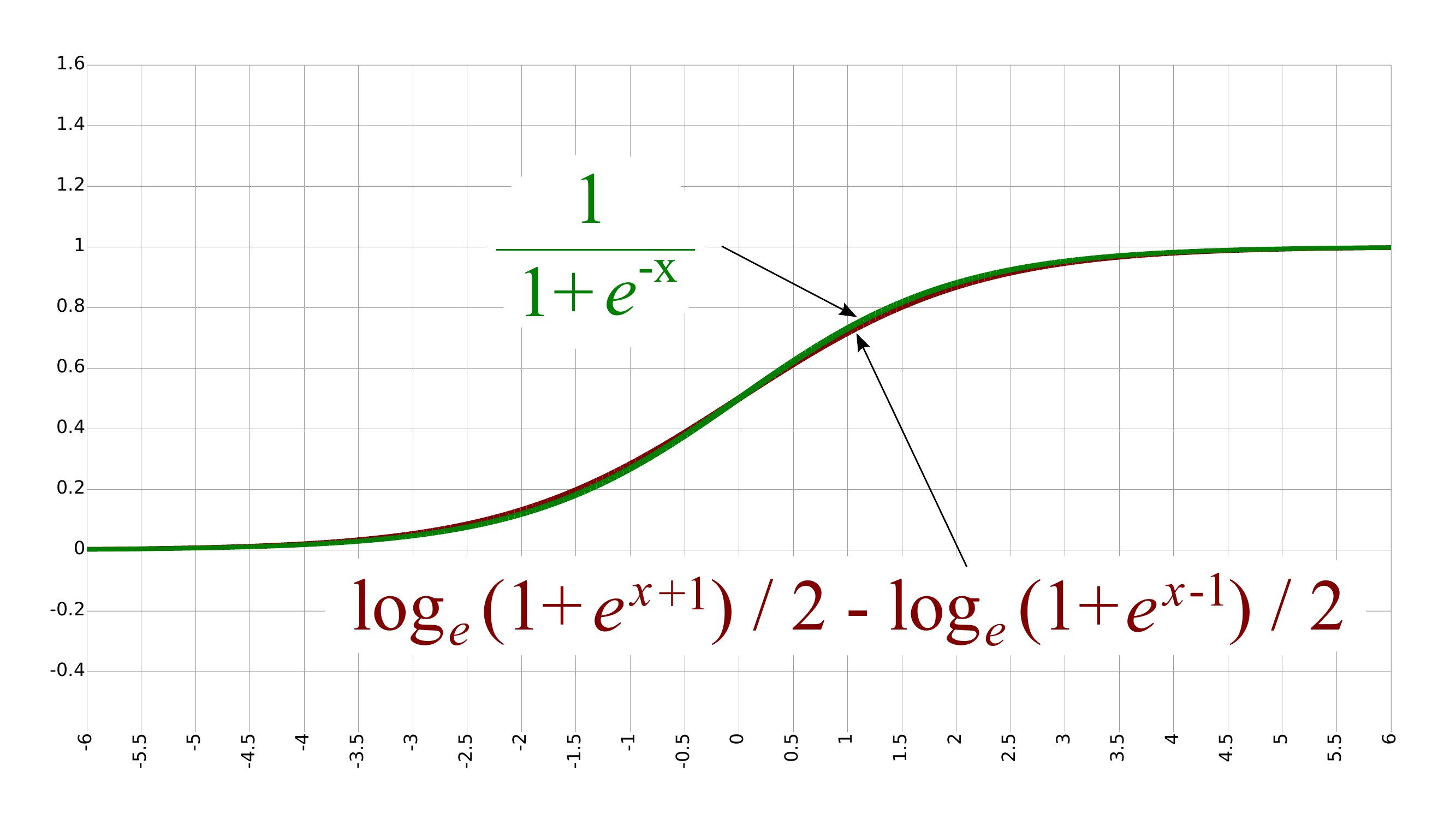}
		\caption{The sum of two softplus units can very closely approximate
				the behavior of one logistic unit. The green curve is the logistic
				function. The red curve is the sum of two softplus units. An even
				closer approximation than is shown
				in this figure could be achieved using fractional parameter values.}
		\label{fig_two_softplus}
	\end{center}
\end{figure}

Conspicuously absent in our design are units with a logistic activation function, which are commonly
used in neural networks. We do not use these in our network because they have a tendency to
produce bends in the model that are not motivated by the training data.
Figure~\ref{fig_superfluous_bends} illustrates this undesirable behavior. Softplus units, therefore,
can be expected to yield better generalizing predictions. Since softplus units are a softened
variant of rectified linear units, this intuition is consistent with published empirical results
showing that rectified linear units can outperform logistic units \cite{nair2010rectified}.
Furthermore, as visualized in Figure~\ref{fig_two_softplus}, two softplus units can be combined to
approximate the behavior of a single logistic unit. Therefore, any training data that can be fitted
with logistic units can also be fitted with twice as many softplus units. Perhaps, one reason
the research community did not recognize the superiority of softplus units earlier is because
of this necessity to use more of them. Thus, in any pair-wise comparisons using the same number of
network units, logistic units may appear to exhibit more flexibility for fitting to nonlinear
data.

\begin{figure}[!tb]
	\begin{center}
		\includegraphics[width=4.7in]{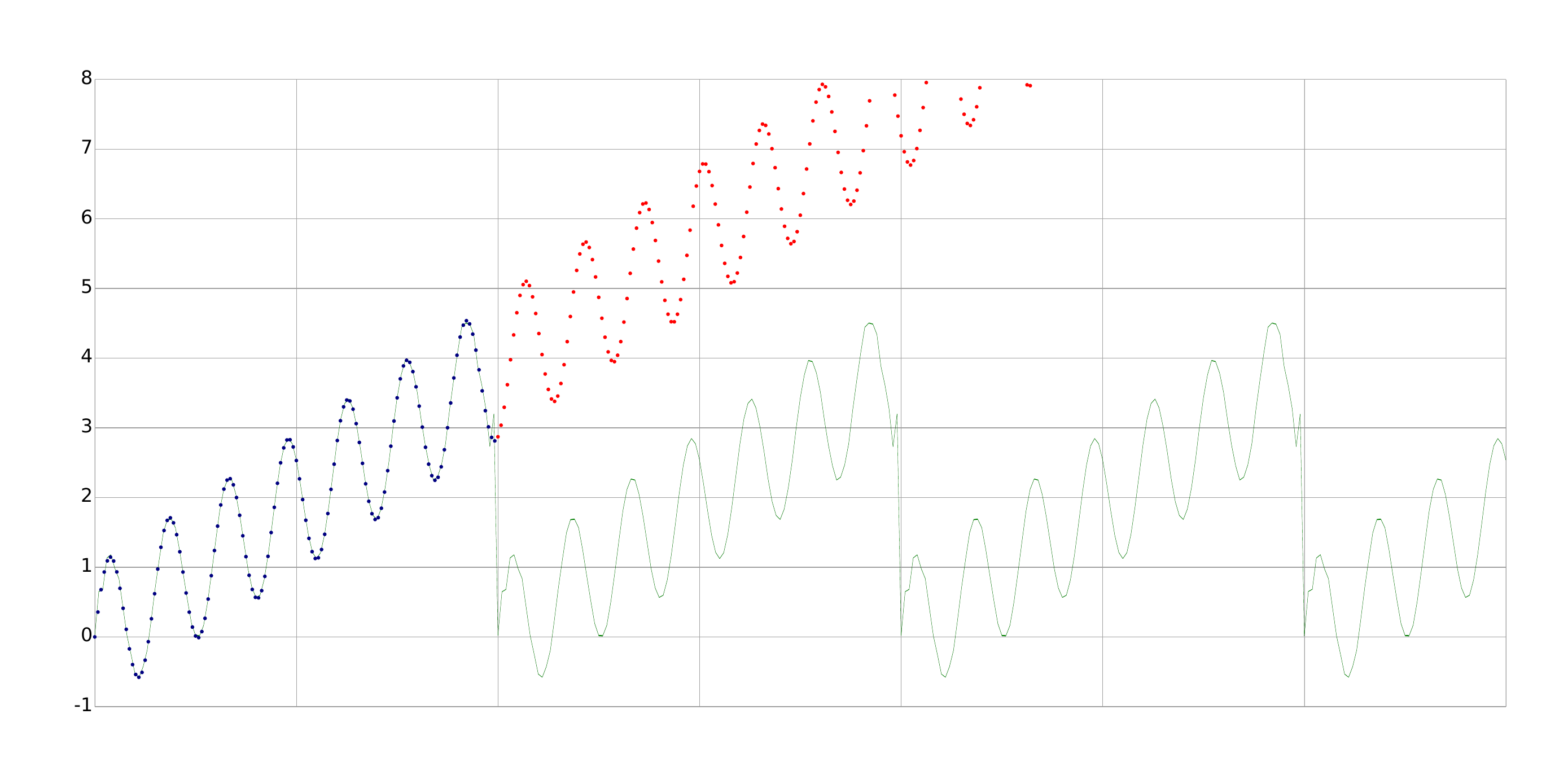}
		\caption{\textbf{Blue points:} Training values generated using the equation $f(t)=\sin(t)+0.1t$.
				\textbf{Green curve:} A model consisting of sinusoid units with weights initialized
				by the fast Fourier transform. \textbf{Red points:} Test values. The fast Fourier
				transform initializes weights in a manner that perfectly fits the training sequence, but
				generalizes very poorly.}
		\label{fig_fourier_init}
	\end{center}
\end{figure}

As training begins, we use the fast Fourier transform to compute weights for the sinusoidal
units, such that they perfectly reconstruct the training signal. (Similar approaches have
been previously proposed \cite{mingo2004fourier,tan2006fourier,zuo2009fourier}.)
Unfortunately, the fast Fourier transform always results in a model that
predicts the training sequence will repeat in the future, as depicted in Figure~\ref{fig_fourier_init}.
Such models tend to generalize very poorly.

\subsection{Regularization}

In order to achieve better nonlinear extrapolation, it is necessary to simplify the model.
Occam's razor suggests that the simplest model capable of fitting to the training data is
most likely to yield the best generalizing predictions. The simplest unit in our model is the
linear unit. Therefore, it is desirable to shift as much of the total weight in
the network away from the sinusoid units and onto the linear units, while still fitting the
model with the training data. Where non-recurring nonlinearities occur in the training data,
linear units will not be able to fit with the training data. In such cases, we want the weight
to shift toward the softplus units. Where repeating nonlinearities occur in the training data,
the sinusoid units should retain some of their initial weight. We accomplish this weight-shifting
during training using a combination of regularization techniques.

A common approach for regularizing a neural network is weight decay, or more specifically $L^2$
regularization. In stochastic gradient descent, $L^2$ regularization is
implemented by multiplying all the weights in the neural network by some factor, $1-\eta\lambda$,
before the presentation of each training pattern. ($\eta$ is the learning rate for training,
and $\lambda$ is a small term that controls how strongly the network is regularized.) $L^2$
regularization prevents the network weights from growing too large during training.
This tends to prevent the curve or surface represented by the neural network from
exhibiting dramatic bends or folds.

A related technique is $L^1$ regularization. It is implemented by subtracting the term,
$1-\eta\lambda$, from all positive weights, and adding that term to all weights that have a negative
value. $L^1$ regularization tends to promote sparsity in the neural network. It causes the
network to utilize as few non-zero weights as possible while fitting the training data.
It tends to be effective when a signal is comprised of a small number of components.

In our experimentation, we found that $L^2$ regularization is much more effective at shifting
weight away from the sinusoid units. Unfortunately, $L^2$ regularization tends to distribute
weight among all of the network units. Since it would be very difficult to select a topology
\emph{a priori} that contains a sufficient number of network units without containing an
excessive number of network units, it is intuitive that the sparse models that result from
$L^1$ regularization may generalize better. Therefore, we begin training with $L^2$ regularization
and complete training with $L^1$ regularization. In our initial experiments, we tried using
$L^p$ regularization, and slowly adjusting the value of $p$ from 2 down to 1. Unfortunately,
computing this exponent is a somewhat expensive operation. Therefore, we also tried the simpler
approach of using $L^2$ regularization during the first half of training, and $L^1$ regularization
during the second half of training. We found this simpler approach to work nearly as well.

In order to bias the network toward shifting weight onto the simpler units, we use a non-uniform
regularization term. For sinusoid units, we regularize with the standard term $1-\eta\lambda$.
For softplus units, we use $1-0.1\eta\lambda$. For linear units, we use $1-0.01\eta\lambda$.
Thus, the strongest regularization is applied to the sinusoid units, while only very weak
regularization is applied to the linear units. These constant factors (1, 0.1, and 0.01) were
selected intuitively. We briefly attempted to optimize them, but in our experiments small variations
tended to have little influence on the final results. Therefore, we proceeded with these
intuitively selected factors.

\subsection{Training procedure}

Stochastic gradient descent relies on being able to update the network weights without
visiting every pattern in the training data. This works well with the logistic activation
function because its derivative outputs a value close to zero everywhere except for input
values close to zero. This gives it a very local region of influence, such that each
pattern presentation will only make significant changes to the few weights that are
relevant for predicting that pattern. The three activation functions we use in our
algorithm, however, have non-local regions of influence. In other words, each pattern
presentation will often significantly affect many or all of the weights in the network.
This has the effect of giving the network a strong tendency to diverge unless a very
small learning rate is used.

In our experiments we found that the values of suitable learning rates varied significantly
as training progressed. Small learning rates, such as $10^{-4}$, would quickly lead to
divergence. Even smaller learning rates, such as $10^{-6}$ would work for a larger number
of training epochs, but would still result in divergence at some point during training.
The only static learning rates that always lead to convergence were so small that training
became impractical. Therefore, the crux of our training algorithm relies on dynamically
adjusting the learning rate $\eta$, and regularization term $\lambda$, during training.

Before training begins, we measure the standard deviation, $\sigma$, of the training data
to serve as a baseline for accuracy. The output layer of our neural network is a single
linear unit. We call this layer 4. The layer that feeds into the output layer, layer 3,
contains several units of each type: sinusoid, softplus, and linear. We use the Fast
Fourier Transform to initialize the sinusoid units, and we initialize the other units
in a manner such that they intially have little influence on the output. (More specific details
are given in Section~\ref{sec_technical}.)

Layers 1 and 2 consist of only softplus and linear units. These layers are initialized to
approximate the identity function, then are slightly perturbed. These layers serve the
important purpose of enabling the model to essentially ``warp time" as needed to fit the
training data with a sparse set of sinusoid units. In other words, these layers enable the
model to find simple repeating patterns, even in real-world data where some of the
oscillations may not occur at precisely regular intervals. Because these layers are further
from the output end of the model, backpropagation will refine them more slowly than the
other layers. This is desirable because warping time in the temporal region of the training
sequence is somewhat of a ``last resort" method for simplifying the model. Ironically,
deep neural networks are often cited for their ability to produce complex models, but
we are actually using them to produce a simpler model. More specifically, we use
the deeper layers to ``explain away" superfluous complexity in the training sequence.
Since the deeper layers contain no sinusoid units, they will specialize only in the temporal
region of the training data, allowing the the sinusoid units that apply everywhere to fit the
data with a simpler model.

After initialization, but before training begins, the entire network
already fits to the training data with a very small root-mean squared error (RMSE).
Usually, it is much smaller smaller than $0.1\sigma$.
At this point, the model simply predicts that the training data repeats itself into the future,
as in Figure~\ref{fig_fourier_init}. As training proceeds, we use stochastic gradient
descent to keep the RMSE near $0.1\sigma$, and regularization to improve generalization.
(Smaller values result in a tighter fit, but longer training times.)
As training completes, the model no longer predicts that the training data will repeat,
but predicts a continuation of the nonlinear trends exhibited in the training data.

It is important that $\eta$ and $\lambda$ be dynamically adjusted using different
mechanisms, so the ratio between them is free to change. After each epoch of training,
we adjust $\lambda$ such that RMSE stays near $0.1\sigma$. When the RMSE is bigger
than $0.1\sigma$, we make $\lambda$ smaller, and when the RMSE is smaller than $0.1\sigma$,
we make $\lambda$ bigger. In contrast with this approach, we make $\eta$ bigger after
each epoch by multiplying it by a constant factor. Eventually, this will always cause
divergence, which can be detected when the RMSE score becomes very large. When that occurs,
we make $\eta$ much smaller, and restore the network weights to the last point with a
reasonable RMSE score. Specific implementation details about this dynamic tuning process are
given in Section~\ref{sec_technical}.

This dynamic parameter tuning process seeks to reduce training time by keeping the learning
rate large while still ensuring convergence. At times, the learning rate will be too large,
which may temporarily have a negative effect on the dynamic process for tuning $\lambda$,
but we have found that divergence tends to happen quickly. Therefore, most of the time,
$\eta$ is within a reasonable range, and the process for tuning $\lambda$ can operate
without any regard to the current value of $\eta$.

\section{Related Works}\label{sec_related}

\begin{figure}[!tb]
	\begin{center}
		\includegraphics[width=4.7in]{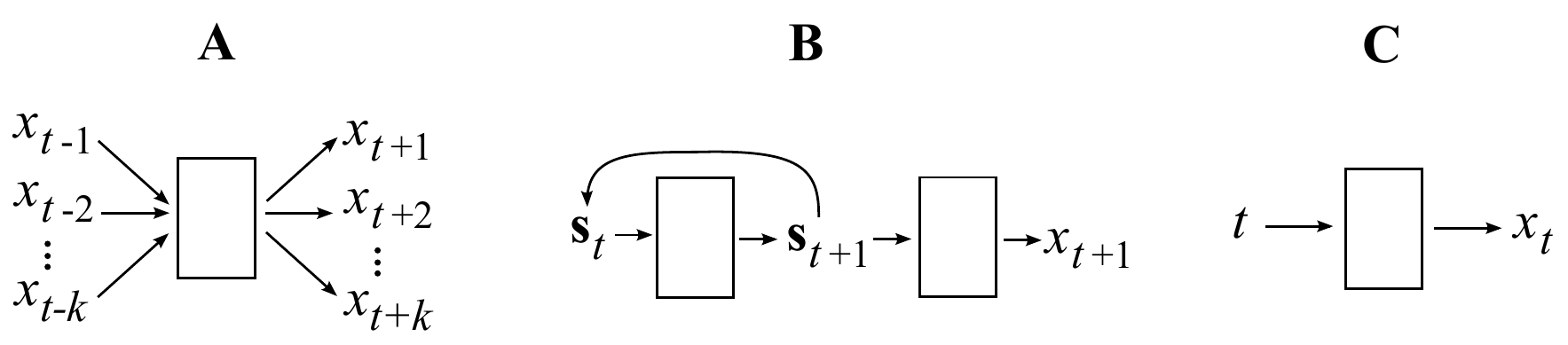}
		\caption{Methods that use neural networks for time-series prediction can be
				broadly grouped into three general approaches. Approach \textbf{A} uses samples from
				the past to predict samples in the future. Approach \textbf{B} uses recurrent connections
				to remember and modify an internal representation of state. Approach \textbf{C} fits a
				curve to the time-series data and uses extrapolation to make predictions. There are
				many variations on each of these approaches. Our work makes advances within group \textbf{C}.}
		\label{fig_types}
	\end{center}
\end{figure}

Many papers have surveyed the various techniques for using neural networks and
related approaches to model and forecast time-series data
\cite{dorffner1996neural,kaastra1996designing,zhang1998forecasting,frank2001time,zhang2003time,de200625}.
Therefore, in this section, we will review only the works necessary to give a high-level overview of
how our method fits among the existing techniques, and defer to these other papers to complete an exhaustive
survey of related techniques. At a high level, the various approaches for training neural networks to
model time-series data may be broadly categorized into three major groups. Figure~\ref{fig_types}
illustrates the basic models that correspond with each of these groups.

The most common, and perhaps simplest, methods for time-series prediction involve feeding sample
values from the past into the model to predict sample values in the future \cite{frank2001time,abarbanel1993analysis}.
(See Figure~\ref{fig_types}.A.) These methods require no recurrent connections, and can be
implemented without the need for any training techniques specifically designed for temporal
data. Consequently, these can be easily implemented using many available machine learning toolkits,
not just those specifically designed for forecasting time-series data. These convenient properties
make these methods appealing for a broad range of applications. Unfortunately, they also have
some significant limitations: The window size for inputs and predictions must be determined
prior to training. Also, they essentially use recent observations to represent state, and they are
particularly vulnerable to noise in the observed values.

A more sophisticated group of methods involves neural networks with recurrent connections
\cite{Nerrand94trainingrecurrent}. These produce their own internal representation
of state. (See Figure~\ref{fig_types}.B.) This enables them to learn how much to adjust their
representations of state based on observed values, and hence operate in a manner more robust
against noisy observations. 

Existing methods for training the weights of recurrent neural networks can be broadly divided into
two categories: Those based on nonlinear global optimization
techniques, such as evolutionary
optimization \cite{floreano:evolve_recurrent_neural_nets,sjoberg:nonlinearblackboxmodeling,blanco:evolutionary_rnn},
and those based on descending a local error gradient, such as Backpropagation Through Time
\cite{mozer1989focused,werbos:bptt} or Real-Time Recurrent Learning
\cite{robinson:rtrl,mozer:rtrl}. 
Unfortunately, in practice, evolutionary optimization tends to be extremely slow, and it is unable
to yield good results with many difficult
problems \cite{sontag:neural_nets_for_control,sjoberg:nonlinearblackboxmodeling}. Gradient-based
methods tend to converge faster than global optimization methods, but they are more susceptible
to problems with local optima. With recurrent neural networks,
local optima are a more significant problem than with regular feed-forward networks
\cite{cuellar:train_recurrent_networks_difficult}. The recurrent feedback can create chaotic
responses in their error surfaces, and can distribute local optima in poor regions of the space.

In early instantiations, recurrent neural networks struggled to retain internal state over long
periods of time because the logistic activation functions typically used with neural networks
tend to diminish the values with each time step. Long Short Term Memory architectures address
this problem by using only linear units with the recurrent loops \cite{hochreiter1997long}.
This advance has made recurrent neural networks much more capable for modeling time-series data.
Recent advances in deep neural network learning have also helped to improve the training of recurrent
neural networks \cite{oh2004gpu,cottrell:new_life_for_neural_nets,gashler:tnldr,graves2013speech}.

The third, and most relevant, group of methods for forecasting time-series data is regression
followed by extrapolation. Extrapolation with linear regression has long been a standard method for
forecasting trends. Extrapolating nonlinear trends, however, has generally been ineffective.
The general model, shown in Figure~\ref{fig_types}.C, is very simple, but training it in a manner
that will make generalizing predictions can be extremely challenging. For this reason, this branch
of time-series forecasting has been much less studied, and remains a relatively immature field.
Our work attempts to jump-start research in this branch by presenting a practical method for
extrapolating nonlinear trends in time-series data.

The idea of using a neural network that can combine basis functions to reconstruct a signal,
and initializing its weights with a Fourier transform, has been previously proposed
\cite{silvescu1999fourier,mingo2004fourier}, and more recently methods for training them
have begun to emerge \cite{tan2006fourier,zuo2009fourier}. These studies, however, do not
address the important practical issues of stability during training and regularizing the model
to promote better generalization. By contrast, our work treats the matter of representing
a neural network that can combine basis functions as a solved problem, and focuses on the
more challenging problem of refining these networks to achieve reliable nonlinear extrapolation.

Many other approaches, besides Fourier neural networks, have been proposed for fitting to
time-series data. Some popular approaches include wavelet
networks \cite{zhang1992wavelet,cao1995predicting,aussem1997combining,geva1998scalenet,subasi2005wavelet,chen2006time},
and support vector machines \cite{huang2005forecasting,lin2006using,sapankevych2009time}.

\section{Technical Algorithm Description}\label{sec_technical}

In this section we describe implementation details necessary to reproduce our results.
The intuition for this algorithm is described in Section~\ref{sec_intuition}.
To assist further research, we have integrated the major parts of our implementation
into the \emph{Waffles} machine learning toolkit \cite{gashler2011jmlr}.

\subsection{Initializing Weights}

Our neural network contains 4 layers. Layer 4 (the output layer) contains a single linear
unit. Layer 3 contains $k$ sinusoid units, where $k$ is a power of 2. In our implementation,
layer 3 also contains 12 softplus units, and 12 linear units. (These quantities can be
adjusted, but due to our use of $L^1$ regularization, there is little need to optimize them.)

We will consider an explanation of the fast Fourier transform to be beyond the scope of this
paper, and will proceed only to describe how to use it to initialize the weights of the sinusoid units.
The fast Fourier transform requires a sequence of $k$ complex values as input. The real component is given by
the values in the training sequence. The imaginary component consists of all zeros. The output of the
fast Fourier transform is also a sequence of $k$ complex values, which we denote as
$\langle \{r_1,i_1\},\{r_2,i_2\},\{r_3,i_3\}, \dots ,\{r_k,i_k\}\rangle$.
Let $w^4_j$ refer to the weight that feeds from sinusoid unit $j$ into the one output unit in layer 4.
Let $b^4$ refer to the input bias of the unit in layer 4.
For each odd value of $j$, $w^4_j$ is initialized to $2r_{(j/2+2)} / k$, where $j/2$ drops the remainder.
For each even value of $j$, $w^4_j$ is initialized to $2i_{(j/2+1)} / k$.
The values in $w^4_k$ and $w^4_{k-1}$ must then be divided by 2.
$b^4$ is initialized to $r_1 / k$.

Let $w^3_j$ refer to the weight that feeds from the first unit in layer 2 into unit $j$ of layer 3.
Let $b^3_j$ refer to the input bias of unit $j$ in layer 3.
For each odd value of $j$, $w^3_j$ is initialized to $2\pi (j/2+1)$, where $j/2$ drops the remainder,
and $b^3_j$ is initialized to $\pi/2$.
For each even value of $j$, $w^3_j$ is initialized to $2\pi (j/2)$,
and $b^3_j$ is initialized to $\pi$.
All other weights feeding into the sigmoid units are initialized to 0.
Note that this parameterization will cause the training values to be represented for input values
between 0 and 1. If the user prefers to represent them from 0 to $k-1$, the weights in layer 3
should be further divided by $k$.

The weights of the linear units are initialized with the identity matrix and their input
bias values are set to zero, such that these units approximate the identity function.
The weights of the softplus units are also initialized to approximate the identity function.
This is done by setting the weights with the identity matrix. The input bias for each
softplus unit is set to a value, $s$. Then, to compensate for this bias, each weight $w_j$ that
feeds out from the softplus unit into the next layer is decremented by $sw_j$. Our
implementation uses the value $s=10$. Larger values will cause the unit to approximate the
identity function more closely, but will cause the unit to require more training to
represent a nonlinear bend in the data.

Layers 1 and 2 each consist of 12 softplus units and 12 linear units. These are initialized
in the same manner as the softplus and linear units in layer 3. Only one input value
(representing time) feeds into layer 1. At this point, the network
should perfectly predict the training sequence, except with some small allowance for the chosen value
of $s$, and the numerical precisions of the various computations. Before training begins, we
slightly perturb all of the weights and input biases by small random values. To do this, we
add random values from a Normal distribution with a mean of 0 and a standard deviation of $10^{-5}$.

To facilitate dynamic parameter tuning, the standard devation of the training sequence
is computed: $\sigma=\sqrt{\frac{1}{k}\sum_i (v_i - \mu)^2}$, where $v_i$ refers to the $i^{th}$
value in the training sequence, and $\mu$ is the mean value in the training sequence.

\subsection{Training}

To begin training, we initialize the learning rate $\eta$ to $10^{-9}$, and the regularization
term $\lambda$ to 1. (Note that these values are dynamically adjusted during training.) 
Each training epoch presents each of the $k$ time-series values in the training sequence to
the network in random order. Each presentation involves feeding a time value in, regularizing
the weights, and then using regular backpropagation to update the weights by stochastic
gradient descent. During the first half of the training epochs, $L^2$ regularization is
used. During the second half of the training epochs, $L^1$ regularization is used.
In each of our experiments, we performed $10^7$ epochs of training.

At the end of each training epoch, the values $\eta$ and $\lambda$ are dynamically adjusted.
This is done by measuring the root mean squared prediction error over the training
values, $\epsilon$. If $\epsilon < 0.1\sigma$, then $\lambda=\lambda * 1.001$ else $\lambda=\lambda / 1.001$.
$\eta$ is always scaled by a factor of 1.01. If $\epsilon < 0.2\sigma$, then a backup
copy of the network weights is made, or else the weights are restored from the most recent backup
copy and $\eta=0.1\eta$. We note that all of these constant values could potentially be optimized,
but we anticipate that only a small amount of performance would actually be gained by doing this,
so we did not actually attempt to optimize them.

\section{Validation}\label{sec_validation}

In this section, we present visual results showing that our method is able to model
nonlinear trends into the future. Deliberately absent in this section are quantitative
comparisons with corresponding results using recurrent neural network methods. If our intent
were to establish our method as the new \emph{state of the art} in time-series forcasting,
then such comparisons would be essential. However, our intent is to present advances in
methods for extrapolation using nonlinear regression. This branch of time-series
forecasting is not yet as well-refined for this task as recurrent neural network methods.
If the research community were to only focus on improving the method with the highest
precision, then it would risk becoming stuck in a local optimum. By advancing this
alternative approach, we intend to help open the way for more research interest in this area,
which could potentially lead to different use-cases or a long-term shift in how time-series
forecasting is done.

\begin{figure}[!tb]
	\begin{center}
		\includegraphics[width=4.7in]{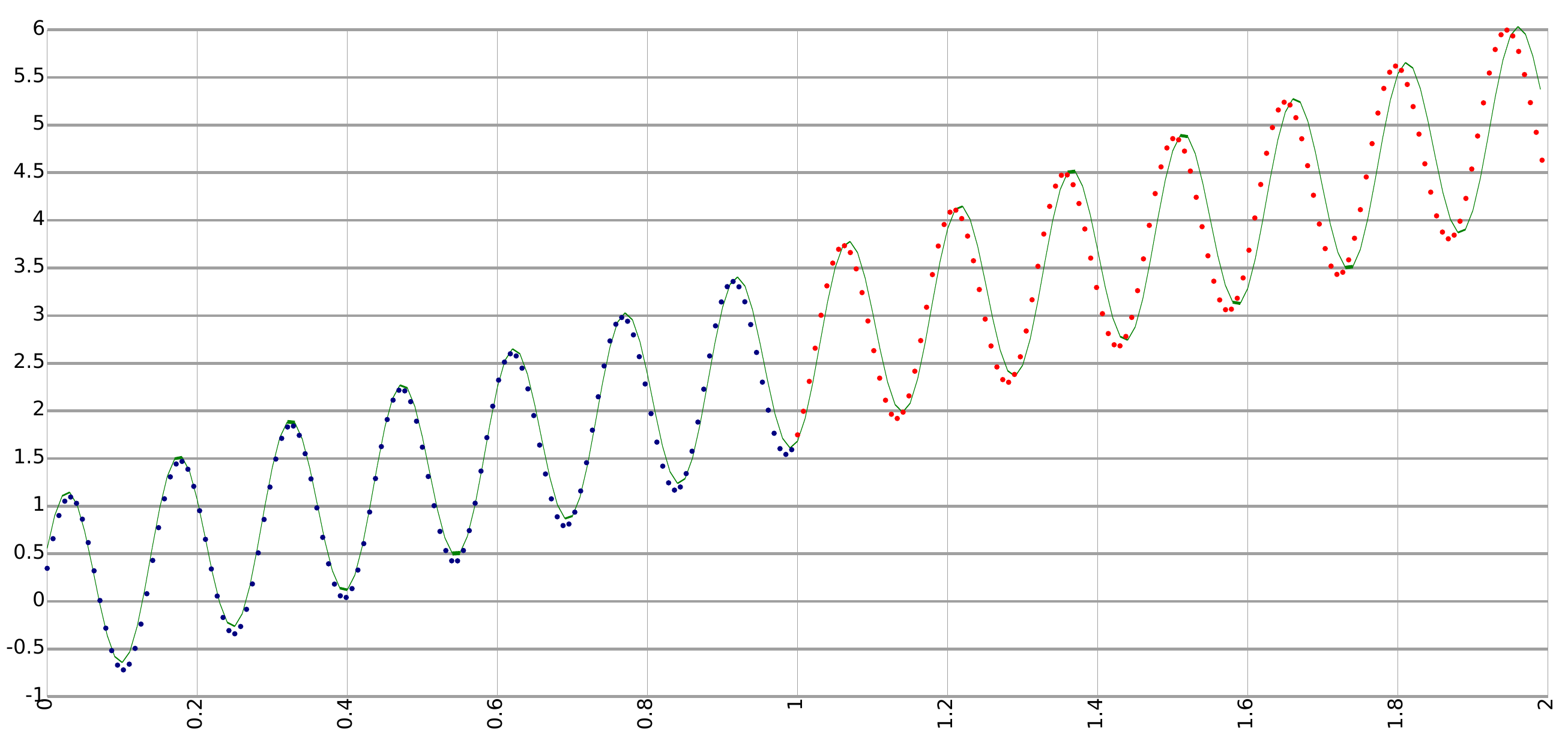}
		\caption{Points generated from the function, $f(t)=\sin(t)+0.1t$. The green line
			represents the continuous model after fitting to the blue points on the left
			half of this chart. The red points on the right half were withheld during
			training so they could serve to evaluate the model.}
		\label{fig_sinplus128}
	\end{center}
\end{figure}

Our first experiment shows results with a toy problem involving a sine wave with the
addition of a linear trend, $f(t)=\sin(t)+0.1t$. For training, we used this equation to
generate a sequence of 128 values. Upon initializing the weights, but before
training, the model predicts that the training sequence repeats into the future, as shown
in Figure~\ref{fig_fourier_init}. This model assigns significant weight to many of its
sinusoid units. As training proceeds, however, the model is greatly
simplified while still fitting with the training sequence. The final model assigns nearly
all of its weight to just two units: a sinusoid unit and a linear unit, which matches the
equation that was used to generate the training sequence. The results are shown in
Figure~\ref{fig_sinplus128}. The blue dots spanning the left half of the plot represent
the training sequence. The red dots spanning the right half of the plot are test values
generated by continuing to sample from the same equation. These were withheld from our
algorithm during training. The green curve shows the continuous predictions of the
trained model.

It can be observed that the model does not perfectly fit the training sequence. This
occurs because we trained it to fit with an RMSE of about $0.1\sigma$. One potential
future improvement to our algorithm might be to decay this value as training progresses,
such that a tighter fit is expected at the end of training. As expected, the model
is slightly less accurate with the test sequence than with the training sequence.
Although the predictions are slightly out of phase, they still model all of the nonlinear
trends in the test sequence very well, which is the problem we attempted to address.

We also evaluated our algorithm on this problem with varying training sequence lengths.
We tested with sizes of 32, 64, 128, 256, and 512. In every case, the results were very
similar, so we only show results for 128 samples as a representative case. These experiments
show that only a few samples are really needed to solve this simple problem, but
using many samples does not cause any problems.

\begin{figure}[!tb]
	\begin{center}
		\includegraphics[width=4.7in]{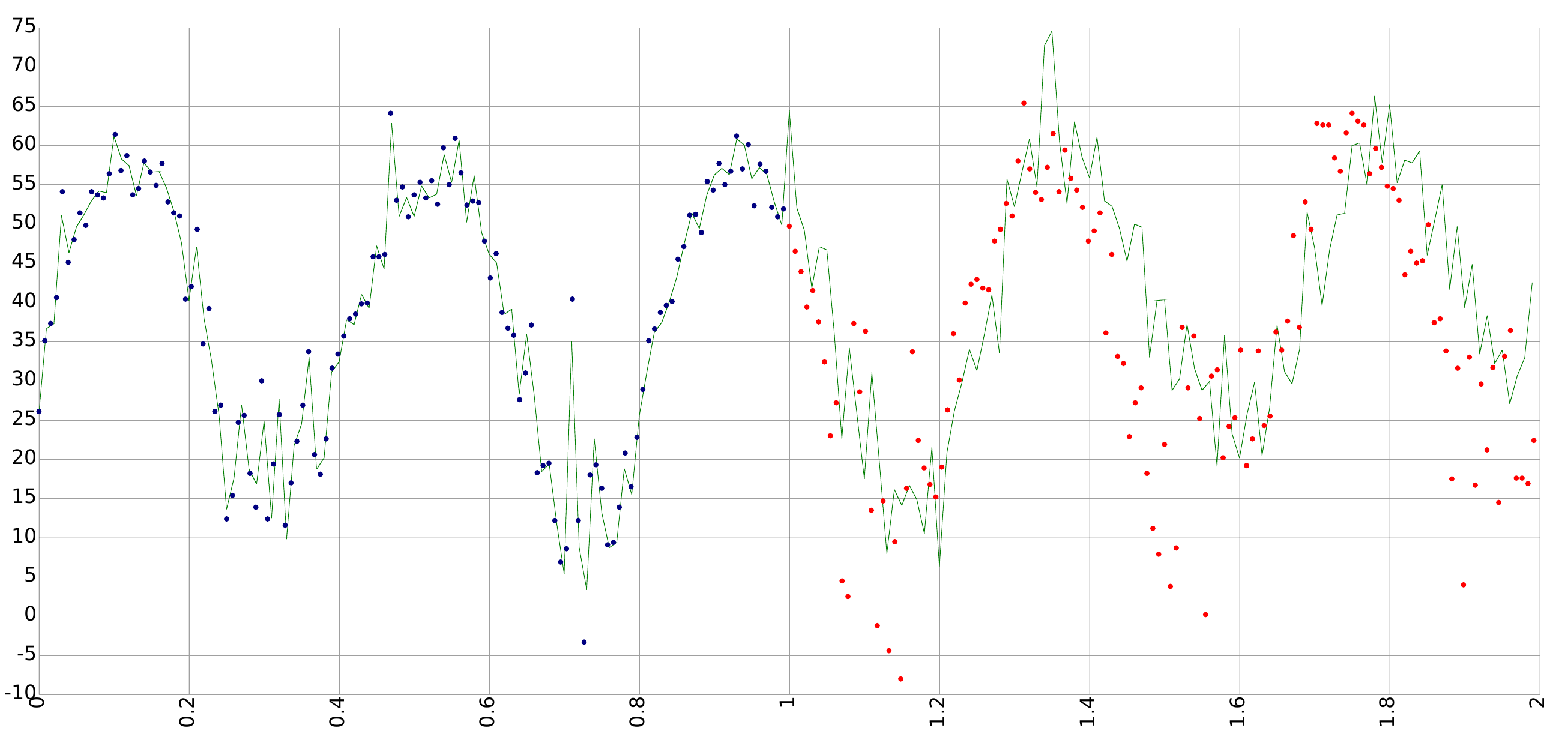}
		\caption{The points represent weekly temperature measurements in Anchorage Alaska
			(in degrees Farenheit) over 5 years beginning in April 2009. The model (shown
			in green) was trained only using the blue (left half) points. It failed to anticipate some
			of the deep temperature plunges in the last three winters, but generally predicted
			the nonlinear temperature trends very well.}
		\label{fig_alaska6}
	\end{center}
\end{figure}

Next, we tested our algorithm with a real-world dataset. We obtained the weekly temperature
measurements in Anchorage Alaska from April 2009 to the present, from http://noaa.gov.
We selected temperature data because it exhibits clear nonlinear trends, and we chose to
use data from Anchorage Alaska because it appeared to fluctuate from a simple sine wave more
than data from other locations that we considered. Our results are shown in Figure~\ref{fig_alaska6}.
The blue points on the left half of the plot were given as the training sequence. The red
points on the right half were used as the test sequence.

\begin{figure}[!tb]
	\begin{center}
		\includegraphics[width=4.7in]{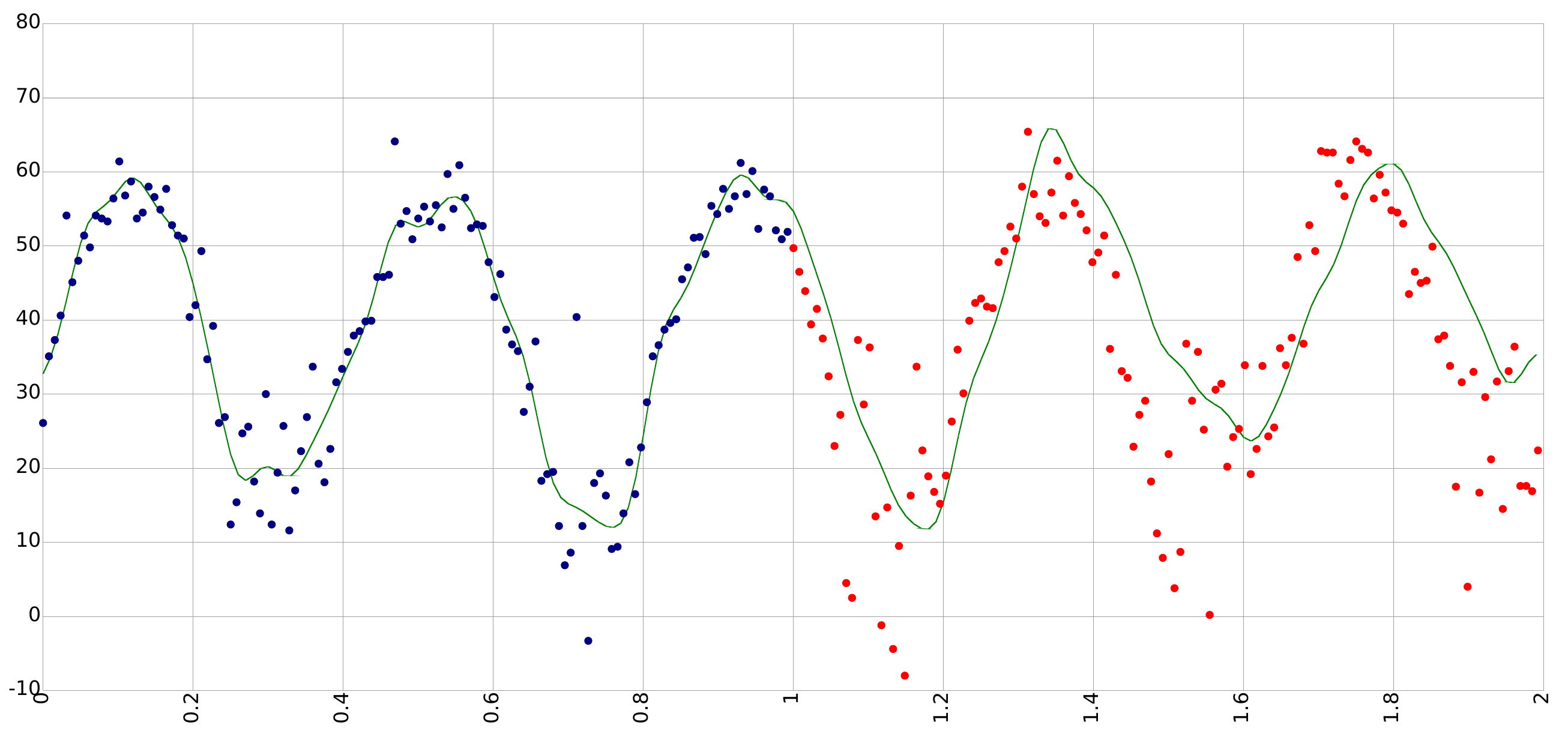}
		\caption{As a post-processing step, we applied a low-pass filter to our model by setting
			the weights feeding from the highest-frequency half of the sinusoid units into the
			linear output unit to zero. The resulting model is able to extrapolate the general
			trends without predicting high-frequency fluctuations. This also makes it more apparent
			how well our model anticipated nonlinear future trends. Whether the high-frequency
			oscillations that this technique removes were entirely noise, or also contained meaningful
			information, remains to be determined.}
		\label{fig_alaska6lowpass}
	\end{center}
\end{figure}

As with the previous experiment, the test results were slightly out of phase. It also failed
to anticipate some of the deep temperature plunges that occurred during the last few winters.
Nevertheless, our extrapolation was able to anticipate the principal nonlinear trends quite well.
This is impressive considering that it was trained on data that spanned fewer than 3 full periods
of the primary oscillating trend in the data. Whether the high-frequency fluctuations predicted by
the model are noise, or contain some meaningful information, is not clear. However, it is simple
to remove them from the model. This is done by setting the weights that feed from the high-frequency
sinusoid units into the linear output unit to zero. Figure~\ref{fig_alaska6lowpass} shows
results after zeroing out the highest-frequency half of the sinusoid nodes as a post-processing
step. The non-uniform shape of the resulting model shows that our model has done more than merely
fit a single sinusoid to the data, and its close fit with the future data shows that it effectively
anticipated nonlinear temperature trends.

\begin{figure}[!tb]
	\begin{center}
		\includegraphics[width=4.7in]{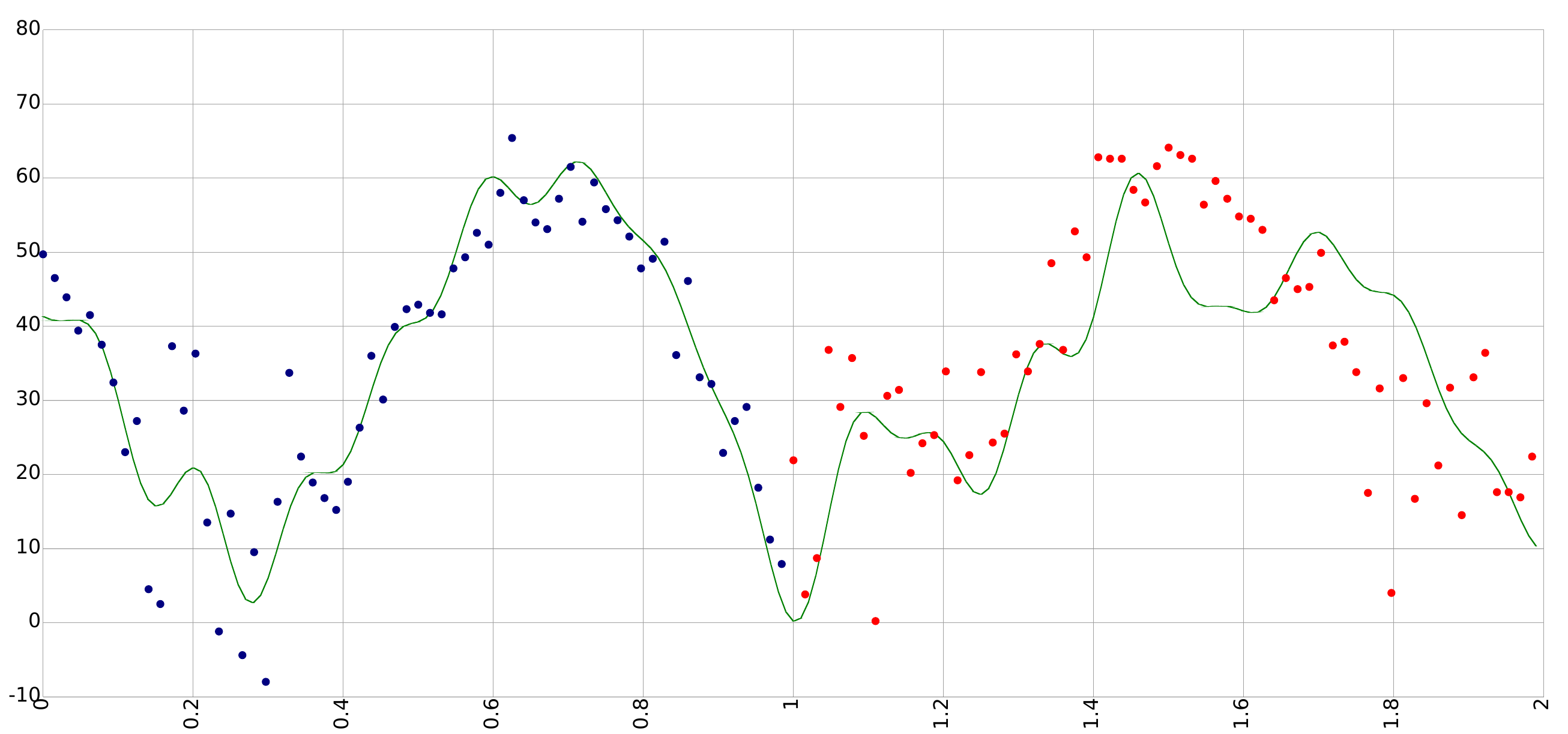}
		\caption{To show that our technique works well even when little training data is available,
			we trained a model using only the first half of the test data from Figure~\ref{fig_alaska6lowpass},
			and tested this model using the second half of the test data. Low-pass filtering was
			also used with these results. Even with so few available training points, the extrapolated
			trend still fit the actual trend reasonably well.}
		\label{fig_alaska7}
	\end{center}
\end{figure}

To demonstrate that our algorithm works well with real-world data when few training points are
available, we repeated this experiment using only 64 training points. Specifically, we divided the
test sequence from the Anchorage temperature problem into a training and test sequence for this problem.
We also applied a low-pass filter to these results, shown in Figure~\ref{fig_alaska7}. Note that
even with so few available training points, the extrapolated trend still fits the actual trend effectively.

\begin{figure}[!tb]
	\begin{center}
		\includegraphics[width=4.7in]{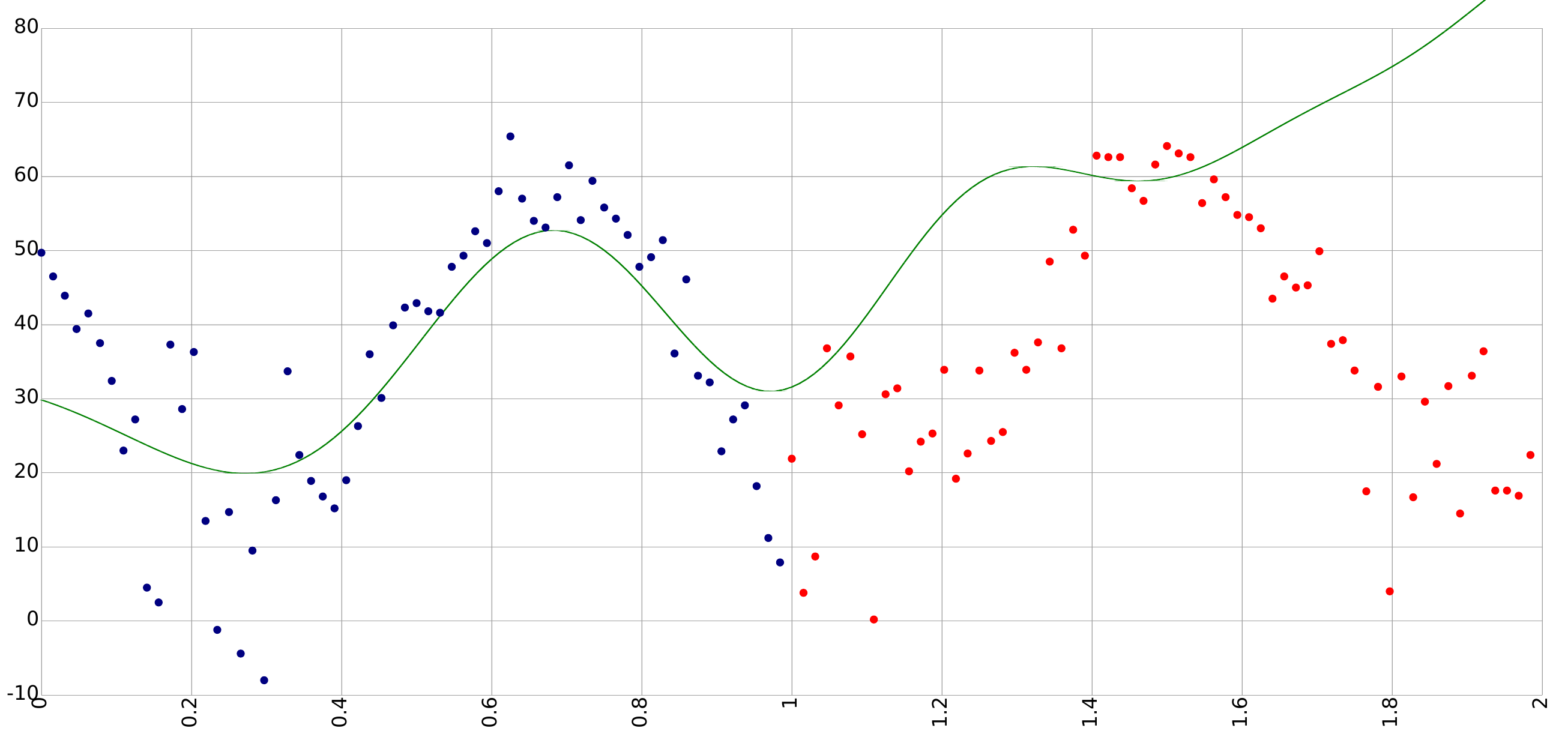}
		\caption{We trained a neural network on the same data as in Figure~\ref{fig_alaska7},
			but did not use the fast Fourier transform to initialize weights. These poor results
			emphasize the importance of this techniques.}
		\label{fig_alaska7simple}
	\end{center}
\end{figure}

To show that our method is what leads to these good results, we repeated the experiment using
regular backpropagation. We used an identical network topology, and trained with regularization,
but we did not use the fast Fourier transform to initialize the weights, and we did not dynamically
adjust the parameters during training. As expected, the results of this approach, shown in
Figure~\ref{fig_alaska7simple} were very poor. No post-processing was performed on these results.
We also performed regular backpropagation with each of the other experiments in this paper, and
obtained even worse results. In most cases, it merely produced a linear trend line because
it became stuck in a local optimum very early in the training process. We also empirically measured
the RMSE scores for both training methods over each problem, and found that our approach was
consistently better. Because the regular approach never produced reasonable extrapolation results, we
do not report these empirical comparison scores.

\begin{figure}[!tb]
	\begin{center}
		\includegraphics[width=4.7in]{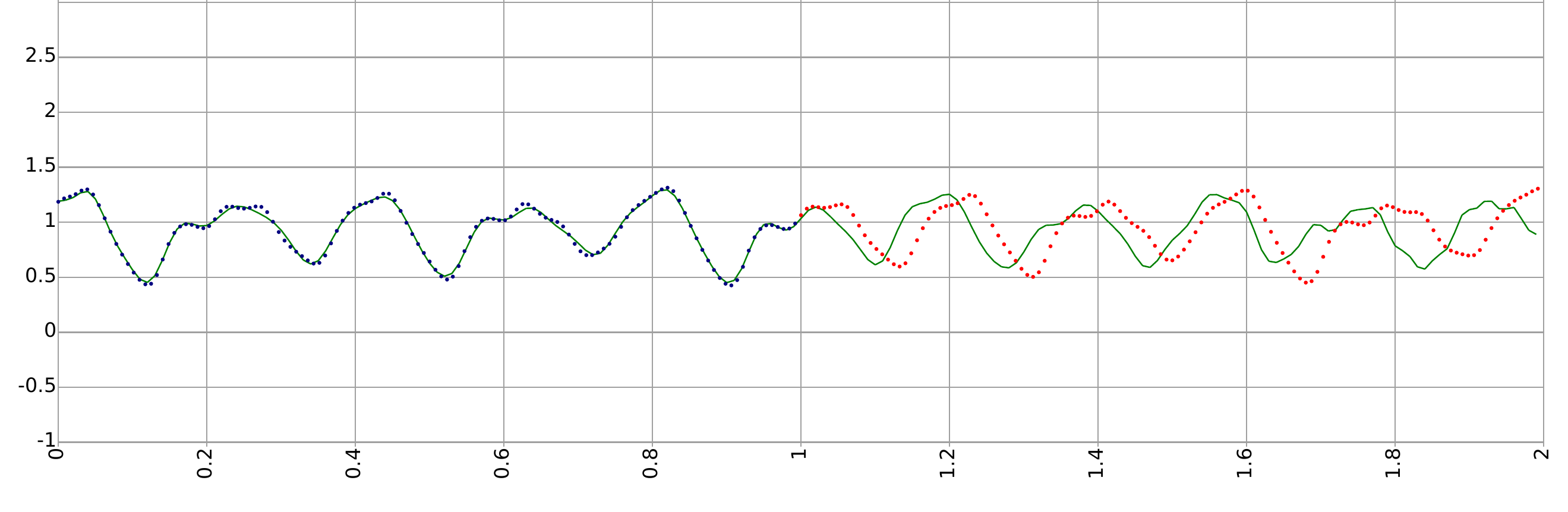}
		\caption{The points are samples from the Mackey-Glass chaotic sequence. The model (shown
			in green) was trained only using the blue (left half) points. Although the predictions
			are slightly out of phase, they accurately anticipate behavior in the data. This is
			significant because these fluctuations do not repeat what occurred in the training
			sequence.}
		\label{fig_mackey14}
	\end{center}
\end{figure}

To demonstrate the versatility of our method, we tested it with the Mackey-Glass series. This test
is interesting because it involves a chaotic series, rather than a periodic series. Results with this
data are given in Figure~\ref{fig_mackey14}. As with previous experiments, the blue points on the left
half of the plot represent the training sequence, the red points on the right half represent the test
sequence, and the green curve is the trained model. We note that the model begins to prematurely
descend very early in the test sequence (at approximately time 1.02), which causes its predictions to
be slightly out of phase for the remainder of the test sequence. Nevertheless, the model clearly
exhibits similar patterns to those in the test set. Significantly, these patterns do not repeat
those in the training sequence, nor does the model repeat its earlier predictions. This shows that
our method can be effective for predicting even non-repeating trends in the near term.

\section{Summary of Contributions}\label{sec_contributions}

We presented a method for fitting a neural network to time-series data for the purpose
of extrapolating nonlinear trends in the data. Our method initializes the weights of sinusoid
units using the fast Fourier transform. In order to promote better generalization, our method
adds several units with simpler activation functions to the network, and then trains in a
manner that enables the weight to shift toward these simpler units, while still fitting to
the training data. We utilize a simple dynamic parameter tuning method to train efficiently
while ensuring that the model always fits the training sequence well. We presented results
with several experiments showing that our method is effective at predicting nonlinear
trends in time-series data. This paper makes several contributions to the current knowlege,
which we itemize here:

\begin{list}{$\bullet$}{}
	\item It proposes new theoretical intuition for why deep neural networks can actually facilitate
			finding simpler predictive models than can be found with shallow networks. Specifically,
			the deeper layers provide a mechanism to ``warp time" in the temporal region of the training
			sequence, allowing subsequent layers to fit it with a sparser set of sinusoid units.
	\item It demonstrates that shifting weight toward simpler units can be promoted during training by
			regularizing the complex units more heavily.
	\item It describes a dynamic method for simultaneously tuning both the learning rate and
			regularization terms.
	\item It shows that dynamic tuning can be an effective solution to the instability problems that
			inherently occur with sinusoidal activation functions.
	\item Most significantly, it unifies all of these techniques into a method for nonlinear
			extrapolation with time-series data, and demonstrates that it is both practical and effective.
\end{list}


%
%
\bibliographystyle{splncs03}
\bibliography{refs}

\begin{thebibliography}{10}
\providecommand{\url}[1]{\texttt{#1}}
\providecommand{\urlprefix}{URL }

\bibitem{abarbanel1993analysis}
Abarbanel, H.D., Brown, R., Sidorowich, J.J., Tsimring, L.S.: The analysis of
  observed chaotic data in physical systems. Reviews of modern physics  65(4),
  1331 (1993)

\bibitem{aussem1997combining}
Aussem, A., Murtagh, F.: Combining neural network forecasts on
  wavelet-transformed time series. Connection Science  9(1),  113--122 (1997)

\bibitem{blanco:evolutionary_rnn}
Blanco, A., Delgado, M., Pegalajar, M.C.: A real-coded genetic algorithm for
  training recurrent neural networks. Neural Networks  14(1),  93--105 (2001)

\bibitem{cao1995predicting}
Cao, L., Hong, Y., Fang, H., He, G.: Predicting chaotic time series with
  wavelet networks. Physica D: Nonlinear Phenomena  85(1),  225--238 (1995)

\bibitem{chen2006time}
Chen, Y., Yang, B., Dong, J.: Time-series prediction using a local linear
  wavelet neural network. Neurocomputing  69(4),  449--465 (2006)

\bibitem{ciresan2011committee}
Ciresan, D., Meier, U., Masci, J., Schmidhuber, J.: A committee of neural
  networks for traffic sign classification. In: Neural Networks (IJCNN), The
  2011 International Joint Conference on. pp. 1918--1921. IEEE (2011)

\bibitem{cottrell:new_life_for_neural_nets}
Cottrell, G.W.: New life for neural networks. Science  313(5786),  454--455
  (2006)

\bibitem{cuellar:train_recurrent_networks_difficult}
{Cu\'{e}llar}, M.P., Delgado, M., Pegalajar, M.C.: An application of non-linear
  programming to train recurrent neural networks in time series prediction
  problems. Enterprise Information Systems VII pp. 95--102 (2006)

\bibitem{cybenko1989ann_universal_function_approximators}
Cybenko, G.: Approximation by superpositions of a sigmoidal function.
  Mathematics of control, signals and systems  2(4),  303--314 (1989)

\bibitem{de200625}
De~Gooijer, J.G., Hyndman, R.J.: 25 years of time series forecasting.
  International journal of forecasting  22(3),  443--473 (2006)

\bibitem{dorffner1996neural}
Dorffner, G.: Neural networks for time series processing. In: Neural Network
  World. Citeseer (1996)

\bibitem{fan2014learning}
Fan, H., Cao, Z., Jiang, Y., Yin, Q., Doudou, C.: Learning deep face
  representation. arXiv preprint arXiv:1403.2802  (2014)

\bibitem{floreano:evolve_recurrent_neural_nets}
Floreano, D., Mondada, F.: Automatic creation of an autonomous agent: Genetic
  evolution of a neural-network driven robot. From animals to animats  3,
  421--430 (1994)

\bibitem{frank2001time}
Frank, R.J., Davey, N., Hunt, S.P.: Time series prediction and neural networks.
  Journal of Intelligent and Robotic Systems  31(1-3),  91--103 (2001)

\bibitem{gashler2011jmlr}
Gashler, M.S.: Waffles: A machine learning toolkit. Journal of Machine Learning
  Research  MLOSS 12,  2383--2387 (July 2011),
  \url{http://www.jmlr.org/papers/volume12/gashler11a/gashler11a.pdf}

\bibitem{gashler:tnldr}
Gashler, M.S., Martinez, T.R.: Temporal nonlinear dimensionality reduction. In:
  Proceedings of the International Joint Conference on Neural Networks. pp.
  1959--1966. IEEE Press (2011)

\bibitem{geva1998scalenet}
Geva, A.B.: Scalenet-multiscale neural-network architecture for time series
  prediction. Neural Networks, IEEE Transactions on  9(6),  1471--1482 (1998)

\bibitem{graves2013speech}
Graves, A., Mohamed, A.r., Hinton, G.: Speech recognition with deep recurrent
  neural networks. In: Acoustics, Speech and Signal Processing (ICASSP), 2013
  IEEE International Conference on. pp. 6645--6649. IEEE (2013)

\bibitem{hochreiter1997long}
Hochreiter, S., Schmidhuber, J.: Long short-term memory. Neural computation
  9(8),  1735--1780 (1997)

\bibitem{huang2005forecasting}
Huang, W., Nakamori, Y., Wang, S.Y.: Forecasting stock market movement
  direction with support vector machine. Computers \& Operations Research
  32(10),  2513--2522 (2005)

\bibitem{kaastra1996designing}
Kaastra, I., Boyd, M.: Designing a neural network for forecasting financial and
  economic time series. Neurocomputing  10(3),  215--236 (1996)

\bibitem{krizhevsky2012imagenet}
Krizhevsky, A., Sutskever, I., Hinton, G.: Imagenet classification with deep
  convolutional neural networks. In: Advances in Neural Information Processing
  Systems 25. pp. 1106--1114 (2012)

\bibitem{le2011building}
Le, Q.V., Ranzato, M., Monga, R., Devin, M., Chen, K., Corrado, G.S., Dean, J.,
  Ng, A.Y.: Building high-level features using large scale unsupervised
  learning. arXiv preprint arXiv:1112.6209  (2011)

\bibitem{lin2006using}
Lin, J.Y., Cheng, C.T., Chau, K.W.: Using support vector machines for long-term
  discharge prediction. Hydrological Sciences Journal  51(4),  599--612 (2006)

\bibitem{mingo2004fourier}
Mingo, L., Aslanyan, L., Castellanos, J., Diaz, M., Riazanov, V.: Fourier
  neural networks: An approach with sinusoidal activation functions.
  International Journal ITA  11(1) (2004)

\bibitem{mozer:rtrl}
Mozer, M.C.: A focused backpropagation algorithm for temporal pattern
  recognition. Backpropagation: theory, architectures, and applications pp.
  137--169 (1995)

\bibitem{mozer1989focused}
Mozer, M.C.: A focused back-propagation algorithm for temporal pattern
  recognition. Complex systems  3(4),  349--381 (1989)

\bibitem{nair2010rectified}
Nair, V., Hinton, G.E.: Rectified linear units improve restricted boltzmann
  machines. In: Proceedings of the 27th International Conference on Machine
  Learning (ICML-10). pp. 807--814 (2010)

\bibitem{Nerrand94trainingrecurrent}
Nerrand, O., Roussel-Ragot, P., Urbani, D., Personnaz, L., Dreyfus, G.:
  Training recurrent neural networks: Why and how ? an illustration in
  dynamical process modeling. (1994)

\bibitem{oh2004gpu}
Oh, K.S., Jung, K.: Gpu implementation of neural networks. Pattern Recognition
  37(6),  1311--1314 (2004)

\bibitem{robinson:rtrl}
Robinson, A.J., Fallside, F.: The utility driven dynamic error propagation
  network. Tech. Rep. CUED/F-INFENG/TR.1, Cambridge University, Engineering
  Department (1987)

\bibitem{sapankevych2009time}
Sapankevych, N.I., Sankar, R.: Time series prediction using support vector
  machines: a survey. Computational Intelligence Magazine, IEEE  4(2),  24--38
  (2009)

\bibitem{silvescu1999fourier}
Silvescu, A.: Fourier neural networks. In: Neural Networks, 1999. IJCNN'99.
  International Joint Conference on. vol.~1, pp. 488--491. IEEE (1999)

\bibitem{sjoberg:nonlinearblackboxmodeling}
Sj\"{o}berg, J., Zhang, Q., Ljung, L., Benveniste, A., Deylon, B., Glorennec,
  P.Y., Hjalmarsson, H., Juditsky, A.: Nonlinear black-box modeling in system
  identification: a unified overview. Automatica  31,  1691--1724 (1995)

\bibitem{sontag:neural_nets_for_control}
Sontag, E.: Neural networks for control. Essays on Control: Perspectives in the
  Theory and its Applications  14,  339--380 (1993)

\bibitem{subasi2005wavelet}
Subasi, A., Alkan, A., Koklukaya, E., Kiymik, M.K.: Wavelet neural network
  classification of eeg signals by using ar model with mle preprocessing.
  Neural Networks  18(7),  985--997 (2005)

\bibitem{taigman2014deepface}
Taigman, Y., Yang, M., Ranzato, M., Wolf, L.: Deepface: Closing the gap to
  human-level performance in face verification. Conference on Computer Vision
  and Pattern Recognition (CVPR), 2014  (2014)

\bibitem{tan2006fourier}
Tan, H.: Fourier neural networks and generalized single hidden layer networks
  in aircraft engine fault diagnostics. Journal of engineering for gas turbines
  and power  128(4),  773--782 (2006)

\bibitem{werbos:bptt}
Werbos, P.J.: Generalization of backpropagation with application to a recurrent
  gas market model. Neural Networks  1(4),  339--356 (1988)

\bibitem{wilson2003general}
Wilson, D.R., Martinez, T.R.: The general inefficiency of batch training for
  gradient descent learning. Neural Networks  16(10),  1429--1451 (2003)

\bibitem{zhang2003time}
Zhang, G.P.: Time series forecasting using a hybrid arima and neural network
  model. Neurocomputing  50,  159--175 (2003)

\bibitem{zhang1998forecasting}
Zhang, G., Eddy~Patuwo, B., Y~Hu, M.: Forecasting with artificial neural
  networks:: The state of the art. International journal of forecasting  14(1),
   35--62 (1998)

\bibitem{zhang1992wavelet}
Zhang, Q., Benveniste, A.: Wavelet networks. Neural Networks, IEEE Transactions
  on  3(6),  889--898 (1992)

\bibitem{zuo2009fourier}
Zuo, W., Zhu, Y., Cai, L.: Fourier-neural-network-based learning control for a
  class of nonlinear systems with flexible components. Neural Networks, IEEE
  Transactions on  20(1),  139--151 (2009)

\end{thebibliography}

\end{document}